\documentclass[conference]{IEEEtran}
\usepackage{widetext}

\usepackage[belowskip=-2pt,aboveskip=0pt]{caption}
\usepackage{cite}
\renewcommand{\baselinestretch}{.96}

\AtBeginDocument{%
 \providecommand\BibTeX{{%
    \normalfont B\kern-0.5em{\scshape i\kern-0.25em b}\kern-0.8em\TeX}}}

\usepackage[normalem]{ulem}
\usepackage{multicol}
\usepackage{amsmath,amsfonts}
\usepackage[ruled, vlined, linesnumbered]{algorithm2e}

\SetCommentSty{mycommfont}
\usepackage{graphicx}
\usepackage{subfigure}
\usepackage{comment}
\usepackage{textcomp}
\usepackage{xcolor} 
\usepackage{tikz}
\usepackage{caption}
\usepackage[shortlabels]{enumitem}
\usepackage{listings}
\usepackage{caption}
\usepackage{lipsum} 
\usepackage{ulem}

\usepackage{hyperref}

\DeclareMathOperator*{\argmax}{arg\,max}

\definecolor{sdr}{rgb}{0.0, 0.65, 0.31}

\begin{document}
\bstctlcite{IEEEexample:BSTcontrol}
%\title{On Balancing Latency and Quality of Multi-view\\ 3D Reconstruction at the Edge}

\title{End-to-End Latency Optimization of Multi-view 3D Reconstruction for Disaster Response}

%\author{
%Xiaojie Zhang, Mingjun Li, Andrew Hilton, Amitangshu Pal, Soumyabrata Dey, Saptarshi Debroy\\
%\small{City University of New York, Clarkson University, Indian Institute of Technology Kanpur} %\\ Email: \textit{xzhang6@gradcenter.cuny.edu, amitangshu.pal@temple.edu, saptarshi.debroy@hunter.cuny.edu}
%}

\author{
    \IEEEauthorblockN{Xiaojie Zhang\IEEEauthorrefmark{1}, Mingjun Li\IEEEauthorrefmark{2}, Andrew Hilton\IEEEauthorrefmark{2}, Amitangshu Pal\IEEEauthorrefmark{3}, Soumyabrata Dey\IEEEauthorrefmark{2}, Saptarshi Debroy\IEEEauthorrefmark{1}}
    \IEEEauthorblockA{\IEEEauthorrefmark{1}Computer Science, City University of New York, New York, NY, USA}
    \IEEEauthorblockA{\IEEEauthorrefmark{2}Computer Science, Clarkson University, Potsdam, NY, USA}
    \IEEEauthorblockA{\IEEEauthorrefmark{3}Computer Science and Engineering, Indian Institute of Technology Kanpur, Kanpur, UP, India}
    %Email: \textit{xzhang6@gradcenter.cuny.edu, amitangshu@cse.iitk.ac.in, saptarshi.debroy@hunter.cuny.edu}}
    }

\maketitle

%{\color{sdr}
\begin{abstract}
%With wider adoption of real-time video analytics applications, edge computing is becoming popular in order to satisfy their strict latency requirements by bringing the compute resources closer to the data site.

In order to plan rapid response during disasters, first responder agencies often adopt `bring your own device' (BYOD) model with inexpensive mobile edge devices (e.g., drones, robots, tablets) for complex video analytics applications, e.g., 3D reconstruction of a disaster scene. 
%satisfy the latency requirements of real-time video analytics applications during disaster response,  . 
Unlike simpler video applications, widely used Multi-view Stereo (MVS) based 3D reconstruction applications (e.g., openMVG/openMVS) are exceedingly time consuming, especially when run on such computationally constrained mobile edge devices. Additionally, reducing the reconstruction latency of such inherently sequential algorithms is challenging as unintelligent, application-agnostic strategies can drastically degrade the reconstruction (i.e., application outcome) quality making them useless.
%With wider use of 3D reconstruction applications for visual computing use cases, Multi-view Stereo (MVS) based reconstruction methods, such as `openMVG/openMVS' 
%are becoming increasingly relevant due to their high reliability. However, %openMVG/openMVS 
%such methods are too time consuming and compute intensive for real-world and latency-sensitive use cases. Additionally, reducing the reconstruction latency of such inherently sequential algorithms is challenging as unintelligent, quick-fix strategies can drastically degrade the reconstruction quality making them useless.
%In this paper, we aim to design a latency optimized openMVG/openMVS pipeline that aims to best balance the user's latency and quality needs by running the pipeline on a collaborative edge system.
In this paper, we aim to design a latency optimized MVS algorithm pipeline, %viz. openMVG/openMVS 
with the objective to best balance the end-to-end latency and reconstruction quality by running the pipeline on a collaborative mobile edge environment. The overall optimization approach is two-pronged where: (a) application optimizations introduce data-level parallelism by splitting the pipeline into high frequency and low frequency reconstruction components and (b) system optimizations incorporate task-level parallelism to the pipelines by running them opportunistically on available resources with {\em online} quality control in order to balance both latency and quality.
% given the current resource availability. 
Our evaluation on a hardware testbed using publicly available datasets shows upto $\sim54$\% reduction in latency with negligible loss ($\sim4-7$\%) in reconstruction quality.       
\end{abstract}


%\section{Problems}
%1. 1.0 - 0.99  number of points
%2. resolution  0.7 - 0.6 “Densify point cloud” 
%3. ******multi-mask, one moving object******
%4. SfM (evaluation) and MVS (evaluation), %similarity and difference

\section{Introduction}
%With the widespread adoption of visual computing use cases, such as autonomous driving~\cite{hane20173d}, robotic surveillance~\cite{lv2017orb}, 3D gaming~\cite{kargas2019using}, smart healthcare~\cite{prokopetc2019towards}, and tactical reconnaissance~\cite{bolick2016study}, need to create {\color{red}reliable} augmented/virtual/mixed reality (AR/VR/XR) environments where real world objects need to be placed in a virtual 3D environment and/or visa versa is becoming essential. 
%{\color{red} Mingjun: please include the references for each of the applications. If you found a survey that talks about all the applications that will be good too.} 
%Thus multi-view 3D reconstruction applications have recently become popular in order to efficiently create such reliable immersive environments. Fig.~\ref{fig:3D-reconstruction} shows one such exemplary use case where.....{\color{red} Soumya: please add the figure and description}

\begin{figure}[t]
    \centering
    \includegraphics[width=.48\textwidth]{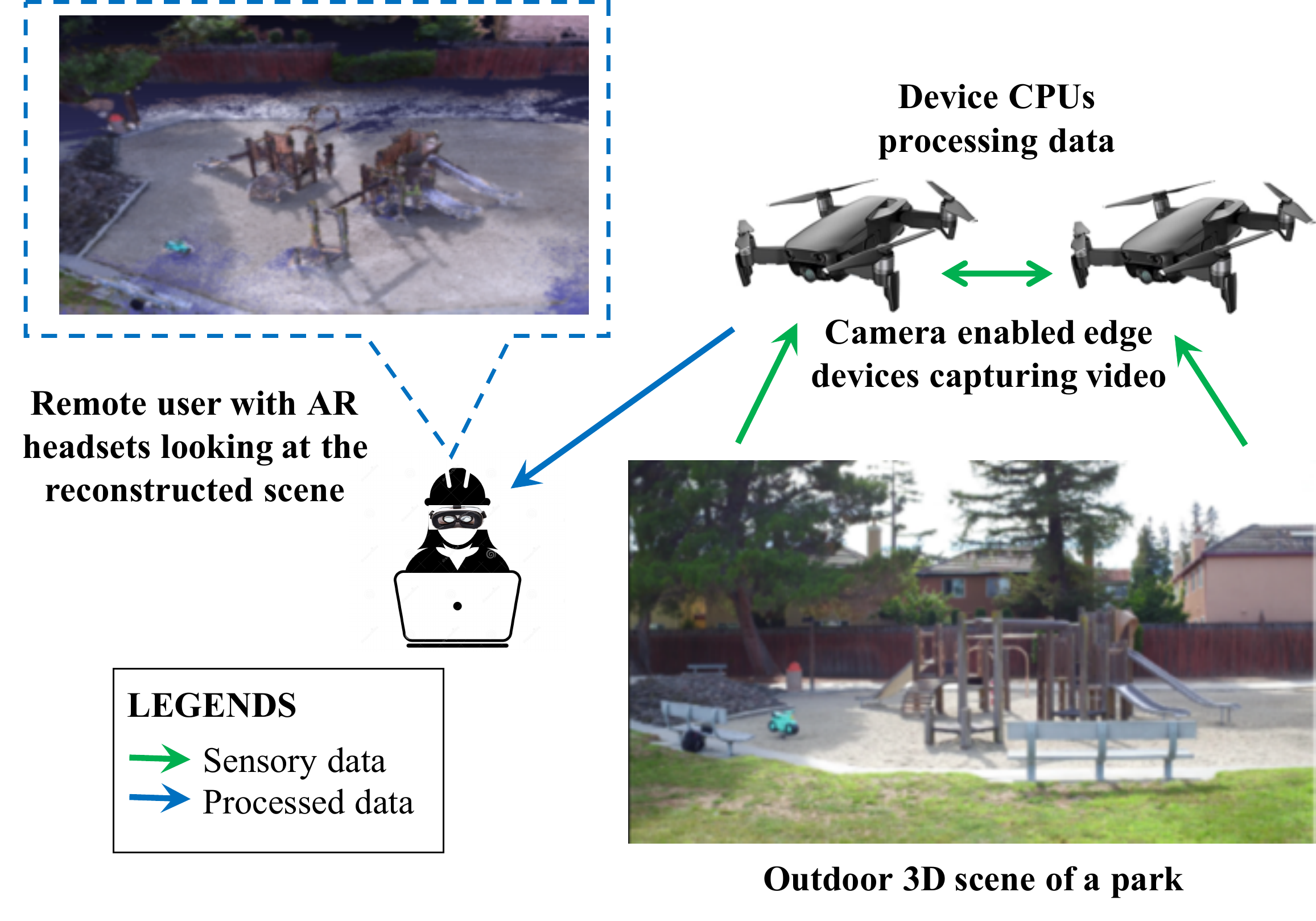}
    \caption{3D reconstruction use case for areal reconstructed view of an outdoor scene using edge devices}
    \label{fig:edge-3D}
\end{figure}

Mobile Edge computing (MEC) in recent times has become an important vehicle and enabler for latency-sensitive video analytics, especially for use cases such as disaster/emergency response~\cite{sers, ccgrid-21}. During disaster response, first responders can use low cost edge devices (e.g., drones, robots, tablets), albeit limited in their computation (e.g., CPU, GPU) capacity for video analytics applications used for rapid situational awareness. Such `bring your own device' (BYOD) based MEC model 
brings compute resources closer to the disaster site when performing computation at distant cloud data centers often becomes expensive and impractical. %{\color{red}
Most of the existing research and adoption of MEC for video analytics have involved simpler applications such as, object detection, object recognition etc~\cite{ccgrid-21}. Here the traditional resource management based latency optimization does not severely impact the analytics outcome due to such applications’ simpler algorithmic structures.
%} 
However, the same can not be said for more complex applications, such as 3D reconstruction that are being increasingly used for disaster response. 
%With the widespread adoption of visual computing use cases, such as 
%autonomous driving~\cite{hane20173d}, robotic surveillance~\cite{lv2017orb}, 3D gaming~\cite{kargas2019using}, smart healthcare~\cite{prokopetc2019towards}, and tactical reconnaissance~\cite{bolick2016study}.
%, 3D reconstruction has become an extremely important problem in order to enable these applications to learn the 3D information of their surrounding environments, both indoors and outdoors. 
%One such indoor application is illustrated in Fig.~\ref{fig:3D-reconstruction}(a) that shows a store robot trying to maneuver around 3D obstacles with the help of multi-camera 3D mapping of store layout and changing positions of the robot. 
%For this the robot uses the two images captured through its eye-cameras to estimate the 3D position of the mug handle.  
%Fig.~\ref{fig:3D-reconstruction}(b) shows an outdoor example where the imagery data obtained from multiple cameras are used to reconstruct the view of a children play area used for surveillance purposes.
%multiple areal drones are used for creating a 3D reconstructed view of a city or a building, which can be used in emergency management scenarios for providing crucial information to the responders.
Fig.~\ref{fig:edge-3D} illustrates a MEC based disaster response where video data obtained from multiple camera-enabled drones are used to reconstruct the view of an outdoor scene used for surveillance purposes. Here the live video data collected from the cameras is processed within these edge devices in a collaborative manner and the processed 3D reconstructed scene is viewed by a user through an  augmented reality (AR) headset.
Traditionally, such 3D reconstruction is achieved by forming geometric relations of the image pixels named epipolar constraints. Many such photogrammetric algorithms, such as Structure from Motion (SfM), compute image features and matchings across views from a set of unordered 2D images~\cite{moulon2016openmvg}. Specifically, SfM is used to generate a sparse 3D point cloud, which is then intensified and textured by Multi-view Stereo (MVS) methods~\cite{openmvs2020}.
%{\color{red}Please check whether yo want to keep Fig. 1(a).}
\begin{comment}
\begin{figure}[t]
    \centering
    \subfigure[] {
        \includegraphics[width=0.985\columnwidth]{3dr_example.png}
        %https://niceclothex.blogspot.com/2021/03/clothing-store-retail-store-layout-3d.html
        \label{fig:a}
    }\\
    \subfigure[] { 
        \includegraphics[width=0.985\columnwidth]{img/3dr_example3.png}
        %https://www.pinterest.com/pin/637400153486839548/
        %https://support.dronedeploy.com/docs/3dmodeling-with-drones
        \label{fig:b}
    }
    \caption{3D reconstruction use cases for (a) robot vision which is relevant for in store mapping in commercial inventories, and (b) for constructing the areal reconstructed view of a backyard.}     
    \label{fig:3D-reconstruction}     
    \vspace*{-18pt}
\end{figure}

\begin{figure}[htb]
    \centering
    \includegraphics[width=.4\textwidth]{img/3D_reconstructionUseCase_3.png}
    \caption{3D reconstruction use case for (a) robot vision which is relevant for in store mapping in commercial inventories, and (b) for the areal reconstructed view of a city/building.}
    \label{fig:edge-3D}
\end{figure}
\end{comment}

\textbf{Latency issues of 3D reconstruction:}
%Among such SfM+MVS pipeline based 3D reconstruction methods, openMVG/openMVS~\cite{moulon2016openmvg,openmvs2020} is one of the top performing platforms that provides high reliability especially due to their excellent performance in reconstructing large scenes~\cite{stathopoulou2019open}.
%\textcolor{cyan}{openMVG/openMVS is also widely popular because of its friendly coding style, modular design, and open-source nature \cite{moulon2016openmvg} making them easy to modify depending on the specific needs of particular applications.}   
%{\color{red} Soumya/Mingjun: please add one/two more lines on the popularity of openMVG+openMVS} {\color{cyan}Both openMVG and openMVS are open source, users can use them for free and make adjustments according to their actual needs}. 
%However, such 
Unlike, simpler visual computing applications, SfM+MVS pipeline based 3D reconstruction methods, such as widely used openMVG/openMVS~\cite{moulon2016openmvg,openmvs2020} are extremely computation-intensive and thereby time-consuming even when performed within an edge environment. This is especially true for reconstructing large real-world scenes (mostly used during response for a emergency scene) that typically need $\geq 3$ high-resolution cameras to capture the target scene from different angles. This is due to the fact that such pipelines are traditionally designed to focus on the reconstruction accuracy and not the processing timeline. %Our initial experiments with openMVG/openMVS pipeline on data from CVSSP-3D Project Sample Datasets~\cite{MustafaICCV15} reveal that single dense 3D reconstruction using 7 camera viewpoints at 100\% resolution takes around $46$ seconds (s) on average. 
However, such high latency of reconstructing a 3D scene is counterproductive for disaster response as many involved missions such as area surveillance and reconnaissance multiple such compute-intensive applications run in sequence.  
%many of the current and future real-world, and often mission-critical edge computing use cases where multiple such compute-intensive applications run in sequence. 
In many of these missions, latency-sensitive applications such as AR/VR/MR, that run after the SfM+MVS based 3D reconstruction pipeline rely on the {\em fast} ($\leq 10$ s) and {\em high quality} reconstruction outcome for the success of the underlying missions (as shown in Fig.~\ref{fig:edge-3D}). 
%One such application can be the store robot (Fig.~\ref{fig:3D-reconstruction}) that needs frequent updates of the 3D store map because of the dynamic nature of the store items. 
Thus, there is a need to optimize such edge-supported 3D reconstruction pipeline's latency without compromising the quality of the reconstruction outcome.
%{\color{red}Saptarshi: please check the below lines and update if necessary.}
%In this paper we propose a latency optimized openMVG+openMVS pipeline on a distributed edge-cloud system that aims to best satisfy the user quality+latency need which is a function over time.     

\textbf{Challenges in balancing latency and quality: }
%{\color{red}
%In this paper, we aim to design a latency optimized openMVG/openMVS pipeline running on an edge system that aims to best satisfy the user expectations in terms of reconstruction latency and quality need which is a function of time.
%} 
However, balancing the trade-off between latency and quality in resource-constrained edge devices is non-trivial as unintelligent application-agnostic quick-fixes can introduce additional challenges. For example:
\begin{itemize}[leftmargin=*,itemsep=0pt]
\item  One possible approach to reduce the 3D reconstruction computation latency running on resource-constrained environments is to separate the dynamic but relatively smaller part (i.e., foreground) of the scene from the static but larger part (i.e., background) and frequently update this foreground to merge it with the less frequent background. However, this is challenging because the SfM pipelines always perform a bundle adjustment optimization, which has an inherent randomness that causes the results to fluctuate so that the foreground and background are located on different coordinate systems affecting the quality of the merged 3D model. Thus there needs to be intelligent application-side optimizations that are tailor made for the reconstruction pipeline.

\item Another way to reduce the latency is to minimize data resolution and use a subset of cameras (i.e., video streams) in order to bring down the total data size for processing. However, on the flip side, unintelligent resolution degradation and camera selection can drastically reduce the 3D reconstruction quality, often rendering it useless. Thus, there is a need for intelligent edge system-side optimization that can balance the pipeline latency and reconstruction quality based on MEC resource availability. 

\item Finally, one of the important prerequisites for such delicate balancing act is the ability to efficiently (in terms of time) and effectively measure reconstruction quality. Quality evaluation needs to be time-efficient so that it can run in parallel to the reconstruction pipeline without eating into the latency savings from application- and system-side optimizations. This in turn makes effective quality evaluation tricky as most state-of-art evaluation techniques~\cite{feng2018evaluation,fan2017point,seitz2006comparison} assume the availability of ground truth which might not be practical when such evaluations need to be lightweight and quick. Contrarily, the absence of ground truth makes the systematic evaluation of 3D reconstruction challenging.
\end{itemize}
{\em Thus overall, in order to effectively perform latency optimizations that do not impact the reconstruction quality, it is of paramount importance that such optimizations are customized towards the 3D reconstruction task pipeline and the underlying algorithms.}

\textbf{Our contributions: }
%The aforementioned challenges towards achieving that can be effectively addressed in three frontiers: a) algorithmic optimization where SfM+MVS pipeline is optimized through opportunistically incorporating AI/ML components that can generate 3D models of the known objects in the scene and perform densify point cloud with low computation cost,b) application/pipeline optimization where the pipeline execution is optimized by opportunistically employing data-level parallelism whenever possible, and c) system optimization where pipelines with instruction-level parallelism are run opportunistically on available edge resources with {\em online} (i.e., `in-the-loop') quality control in order to balance both reconstruction latency and quality given the current availability of edge resources. In this paper, we propose a latency optimized openMVG/openMVS pipeline towards 3D reconstruction that employs the latter two types of optimizations (i.e., application and system) in order to balance pipeline latency and reconstruction quality based on edge resource availability. Particularly:
%\begin{itemize}[leftmargin=*,itemsep=0pt]
In order to address the aforementioned challenges, we propose a latency optimized multi-view 3D reconstruction framework running on a collaborative MEC environment for disaster response that aims to balance the processing time and reconstruction quality. Particularly:
%1) We separate and parallelize the reconstruction of foreground and background areas to reduce the processing complexity. 2) To study the trade-off between processing time and reconstruction quality, we develop lightweight algorithms to adjust the frames' resolution and selection of cameras that satisfies user expectations. 3) We evaluate the performance of our optimized pipeline through experiments on real-world datasets. Our results show that the proposed reconstruction pipeline together with the inter-edge collaboration architecture can save the end-to-end processing time by $61.76\%$ and $70.02\%$ for datasets Dance1 and Odzemok respectively. We also conclude that the quality degradation caused by our proposed pipeline is $5- 8$\%. %The results demonstrate high adaptability and efficiency in optimizing the system objective. 
%The proposed online algorithm can successfully handle different user's deadline requirements based on the availability of edge resources.
%\vspace{-0.05in}
\begin{itemize}[leftmargin=*,itemsep=0pt]
\item In order to reduce processing complexity and thereby latency we introduce data-level parallelism by modifying the SfM+MVS pipeline to create separate pipelines for high frequency foreground reconstruction and low frequency background reconstruction. Through this, we avoided the redundant computation of the static part of the data and ensure MEC resource and time saving.  
%Through such background subtraction we avoid reconstruction on duplicate information.

    %\item {\color{red} Data level parallelism??}
\item We propose a lightweight MEC system optimization algorithm that can select the best configuration of reconstruction latency and quality (that satisfies user expectations) based on resource availability by adjusting the frames' resolution and selection of cameras. 

%{\color{red} It seems like point 3 and 4 are subparts of point 2. Also the last part of point 3 seems redundant.}
\item  %{\color{red}****} 
As part of the MEC system-side optimizations, we propose a novel and lightweight camera selection algorithm in order to select an optimal set of cameras that best covers the target scene. In addition, the proposed algorithm can return a different set based on the desired number of cameras for optimization. 
%This algorithm greatly reduces the searching overhead of the optimal configuration.

\item %{\color{red}****} 
We also propose an {\em online} reconstruction quality evaluation model 
%inspired from F-Score metric~\cite{ruanobenchmark,seitz2006comparison} that can address both the accuracy and completeness of the reconstructed mesh in the absence of ground truth. The {\em online} quality prediction model 
that along with optimal configuration selection, and camera selection algorithms run in parallel to the optimized SfM+MVS pipeline as part of a novel inter-edge collaboration architecture implementing task-level parallelism.  
%\end{itemize}

\item We evaluate the performance of our optimized framework on a hardware testbed with publicly available Dance1 and Odzemok from CVSSP-3D Project Sample Datasets~\cite{MustafaICCV15}. Our results show that the proposed optimized 
framework
%reconstruction pipeline together with the inter-edge collaboration architecture 
can save the end-to-end processing time by $54\%$ for data sets Dance1 and Odzemok. We also conclude that the quality degradation caused by such parallelism is $4- 7$\%. The results demonstrate high adaptability and efficiency in optimizing the system objective. The proposed online algorithm can successfully handle different user requirements during a disaster response mission based on the availability of the MEC resources.
\end{itemize}

\textbf{Paper organization: }
%The rest of the paper is organized as follows.  
In rest of the paper, Section~\ref{Sec:related-work} discusses the related work and background on 3D reconstruction technique and openMVG/openMVS pipeline. Section~\ref{Sec:system-model} presents the system model and benchmarking experiments. Section~\ref{Sec:optimization} discusses the application and system  optimizations. Section~\ref{Sec:experiments} presents the evaluation method and experimental results.  Section~\ref{Sec:conclusions} concludes the paper.

\section{Related Work}
\label{Sec:related-work}

\textbf{Accelerated 3D reconstruction: }% is a technique that applies a series of complex algorithms to a set of images that are taken at different viewpoints for the same scene, and then without any extra information, outputs a corresponding 3D model of that scene. As we discussed in the Introduction, most of the pipelines are very time-consuming because of the need to ensure the quality of reconstruction. In order to reduce the processing time, some researchers have been done their efforts. 
There are a few methods that attempted to optimize the latency of the aforementioned 3D reconstruction pipeline. %In~\cite{fu2018texture}, the authors directly use a RGB-D camera to obtain the depth information and thus avoiding the cumbersome depth map computation algorithm. This process saves a lot of processing time but it requires specialized hardware. Also, the limitation of depth cameras is that they are not very good in outdoor conditions and when objects are far from the camera. 
Authors in~\cite{lou2014cost} sort the images based on the spatial orders of the cameras ensuring large overlaps between two subsequent images of the ordered set. This helps to reduce the computation cost in the feature matching step. %Additionally, they used GPU to parallelize the feature detection steps. However, the optimization performed is in the SfM pipeline and not in the 3D surface with texture generation steps which are the most time-consuming.
In \cite{wang2014improved}, the authors optimize the densify point cloud step with a quasi-dense feature matching approach and achieved 9\% improvement in latency. Authors in~\cite{wang2018fast} group the sparse 3D points into different clusters and processes each cluster separately for dense textured mesh generation. This helps them to reduce around 13\% of the total processing time. %Also, they completely skipped the bundle adjustment optimization which can be detrimental for complex 3D scene reconstruction. 
Compared to these, our work intelligently divides the data into foreground and background regions to significantly reduce the processing time.

\vspace{1mm}
\textbf{Video  analytics:} %Video analytics is becoming an increasingly popular application. The literature on video analytics can be broadly subdivided into two categories: {\em configuration adaptation} and {\em task placement}. 
The performance (e.g., accuracy) of video analytics depends on the joint impact of various configurations. In~\cite{wang2020joint}, the authors propose a framework for both configuration adaptation and bandwidth allocation of edge-assisted video analytics. Authors in~\cite{ran2018deepdecision} jointly consider the interaction between accuracy, video quality, battery constraints, and 
network parameters to find an optimal offloading strategy. 
%In addition to configuration adaptation, placing tasks in the correct computing nodes can impact the overall performance greatly.
VideoStrom~\cite{zhang2017live} optimizes query scheduling by exploring utility-based resource management in terms of query accuracy and delay; while VideoEdge~\cite{8567661} extends the problem to query placement across a hierarchy of clusters. Nevertheless, none of those works can be applied to the multi-view 3D reconstruction applications which is our focus in this paper.

%\vspace{1mm}
%\textbf{Application Profiling:} Application profiling is the basis for optimizing the system, especially for video analytics applications. The profiling result can be fed back to the optimization problem in decision-making. For example in~\cite{yi2017lavea} the authors have characterized the OpenALPR (a plate recognition application) with various resolutions on different computing nodes. Authors in~\cite{zhang2019hetero} have evaluated the relationship between task delay and CPU utilization of a 3D reconstruction application under different degrees of data parallelism as well as various image resolutions. The authors in~\cite{liu2018edge} have studied the latency-accuracy tradeoff of the application and use analytical models to show the relationship between CNN model, frame resolution, and frame rate. However, the above mentioned works use empirical measurements to obtain the application profiling only at the beginning of the execution, which are not feasible in handling the dynamic nature of real-world scenarios. In addition, it is not always practical to assume that the server has perfect information about the application.

\begin{figure}[t!]
%\vspace*{-3pt}
    \centering
    \includegraphics[width=.5\textwidth]{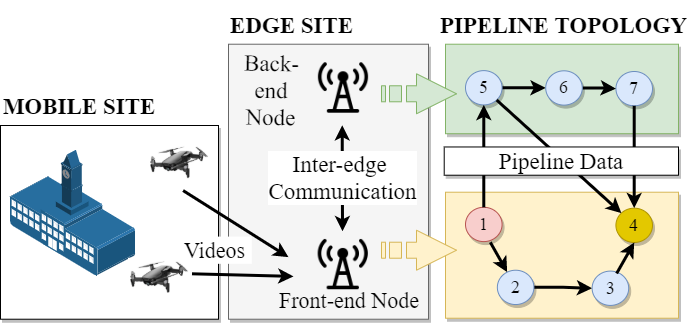}
    \caption{Disaster response system model of collaborative MEC architecture for multi-view 3D reconstruction pipelines}
    \label{fig:system_model}
    %\vspace*{-18pt}
\end{figure}

%\vspace{-3pt}
\section{System Model and Evidence Analysis}
\label{Sec:system-model}

\subsection{System Model and Objectives}

%{\color{blue}
Fig.~\ref{fig:system_model} shows the disaster response system model of our proposed  collaborative MEC architecture implementing the openMVG/openMVS pipeline for end-to-end 3D reconstruction. Our system includes: 1) a group of synchronous video sequences of a scene offloaded by camera enabled edge devices, 2) a front-end edge device/node where the uploaded frames are initially stored and the main 3D reconstruction procedures are undergoing, 3) a back-end edge device/node (helper) which is executing the pipeline in parallel. The edge nodes are connected to each other through point to point wireless or wired link similar to typical real-world setups. % Therefore the data transmission overhead is negligible compared to the computation time. 
%Our system includes a representative user site for data acquisition and an edge site for data computation. The user site has a collection of cameras that send 2D images to the edge site at a particular frequency. This user site can also execute another application sending pre-captured images of a scene from different angles. 
%The edge site contains a front-end edge node where all the uploaded data are initially stored. The communication from the user site to the edge site is wireless. The front-end node is connected to another back-end node through high speed inter-edge communication, i.e.,  dedicated proprietary network. Therefore the data transmission overhead is negligible compared to the computation time. 
In this paper, a video analytics task is defined as reconstructing a dense point cloud from the images captured at the same timestamp. We denote $\mathcal{I}=\{1,2,3...I\}$ as the set of reconstruction tasks for all the timestamps in offloaded videos. We assume that the user has an end-to-end processing deadline requirement %{\color{green}Not clear about this line} 
and requires the best possible reconstruction quality, both calculated as an average of multiple frames. Based on such real-world use case assumptions, the objective of the MEC system is to maximize the average reconstruction quality while meeting the average processing deadline.

\begin{table*}[htb]
\caption{Processing latency and quality of openMVG/openMVS pipeline with varying camera resolution and camera selection}
\label{quality}
\centering
\begin{tabular}{|c||c|c|c|c|c|c|}
\hline
\bfseries Metrics & \bfseries
\begin{tabular}{@{}c@{}}Scale = 1,\\ all cameras\end{tabular}
& \bfseries 
\begin{tabular}{@{}c@{}}Scale = 0.8,\\ all cameras\end{tabular}
& \bfseries 
\begin{tabular}{@{}c@{}}Scale = 0.6,\\ all cameras\end{tabular}
& \bfseries 
\begin{tabular}{@{}c@{}}Scale = 1,\\ sans camera \#0\end{tabular}
& \bfseries 
\begin{tabular}{@{}c@{}}Scale = 1,\\ sans camera \#2\end{tabular}
\\
\hline\hline
Total latency &  26.17 s.  & 16.07 s  & 9.1 s. & 20.98 s. & 19.83 s.\\
\hline
%\begin{tabular}{@{}c@{}}\textbf{Number of Vertices}\\(Sparse Point Cloud)  \end{tabular} &  1486  & 1050  & 663 & 1321 & 1088\\
%\hline
\begin{tabular}{@{}c@{}}Number of Vertices
%\\( in dense point cloud)  
\end{tabular} &  277,790  & 173.139  & 103,493  & 238,821 & 206,513\\
\hline
\end{tabular}
%\vspace*{-12pt}
\end{table*}

\begin{figure*}[htb]
    \centering
    \subfigure[With scale=1, all cameras] {
        \includegraphics[width=0.6\columnwidth]{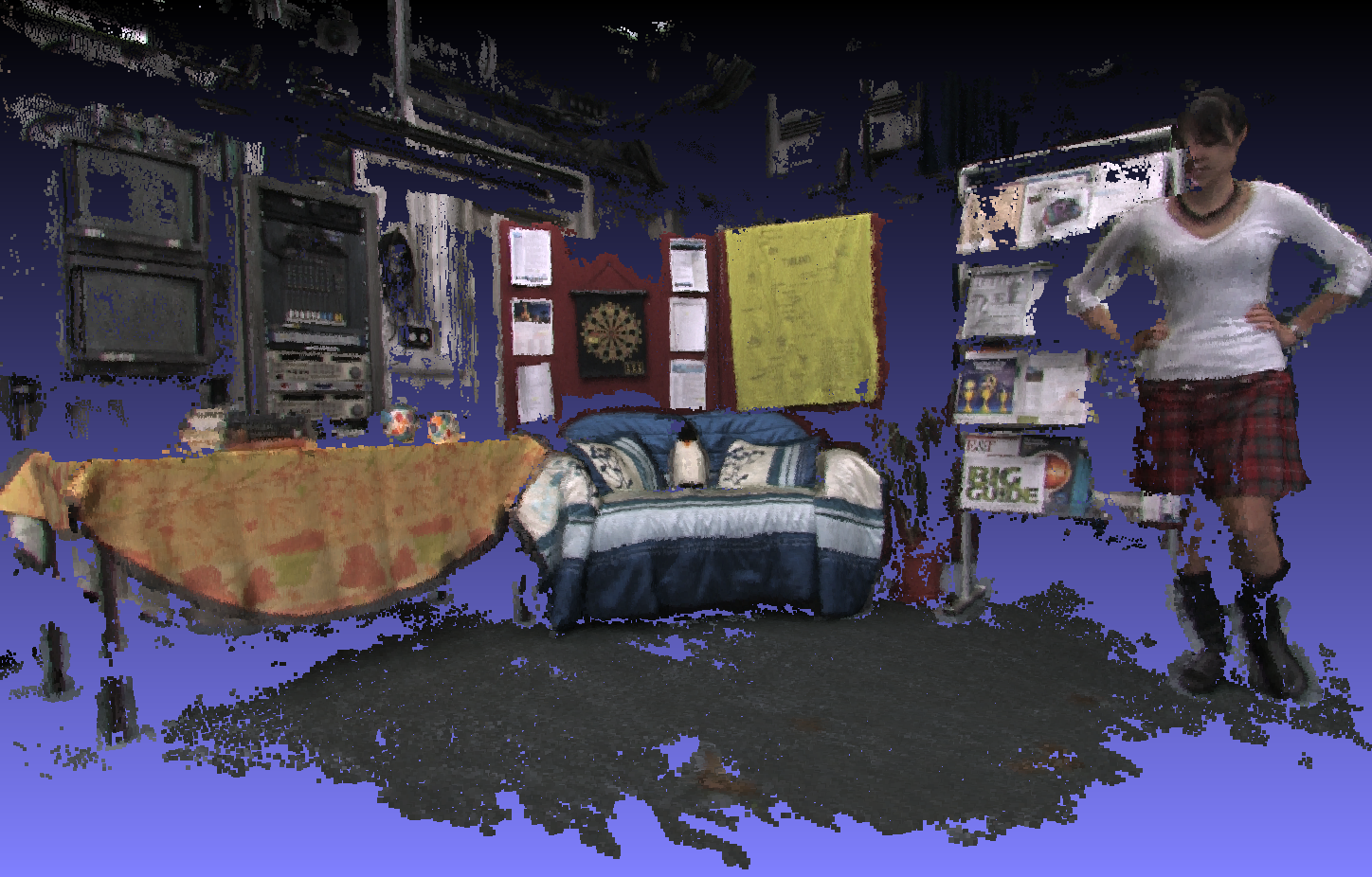}
        \label{fig:a}
    }
    \subfigure[With scale=0.6, all cameras] { 
        \includegraphics[width=0.6\columnwidth]{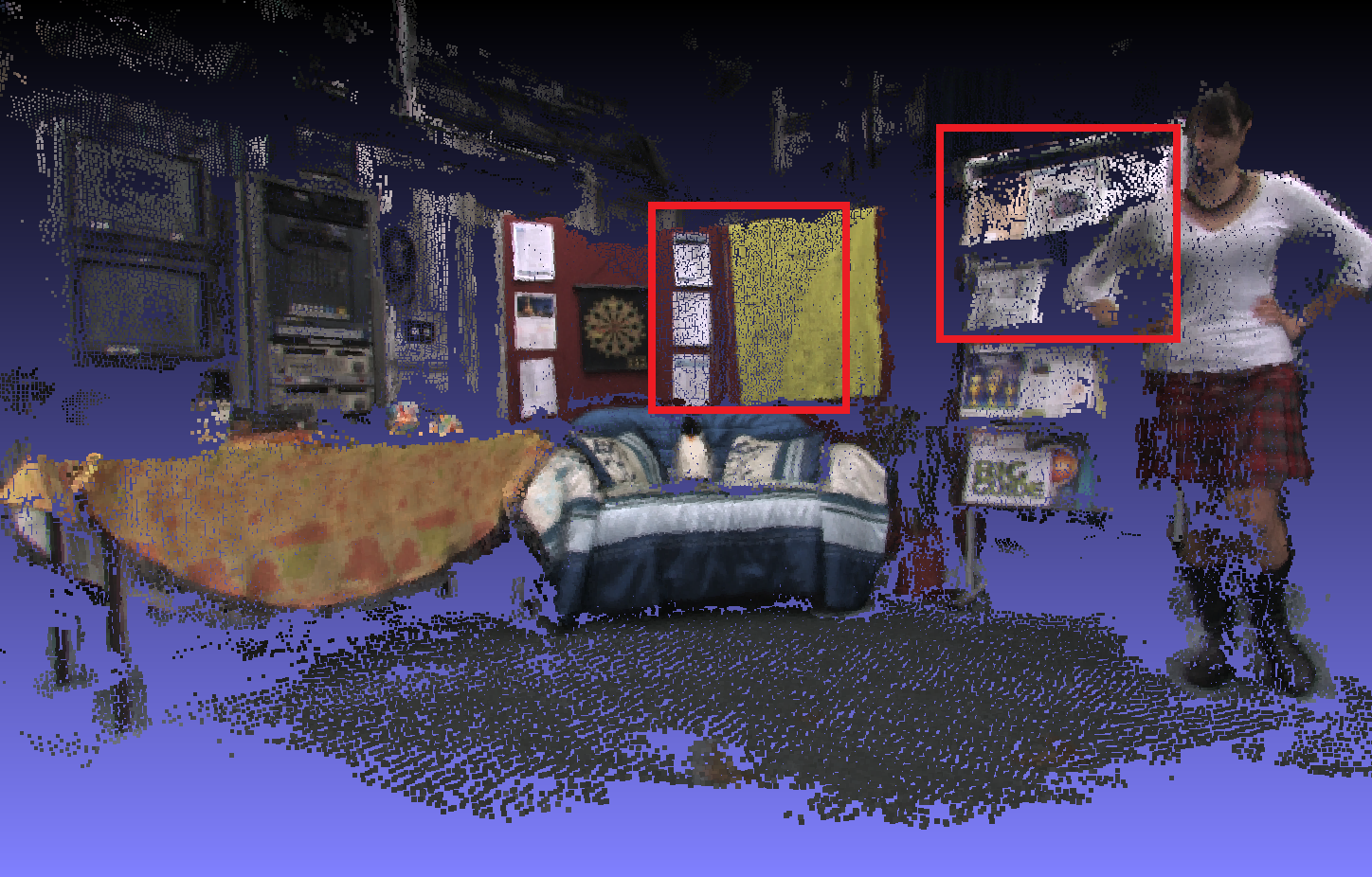}
        \label{fig:b}
    }
    \subfigure[Without scale=1, sans camera \#2] { 
        \includegraphics[width=0.6\columnwidth]{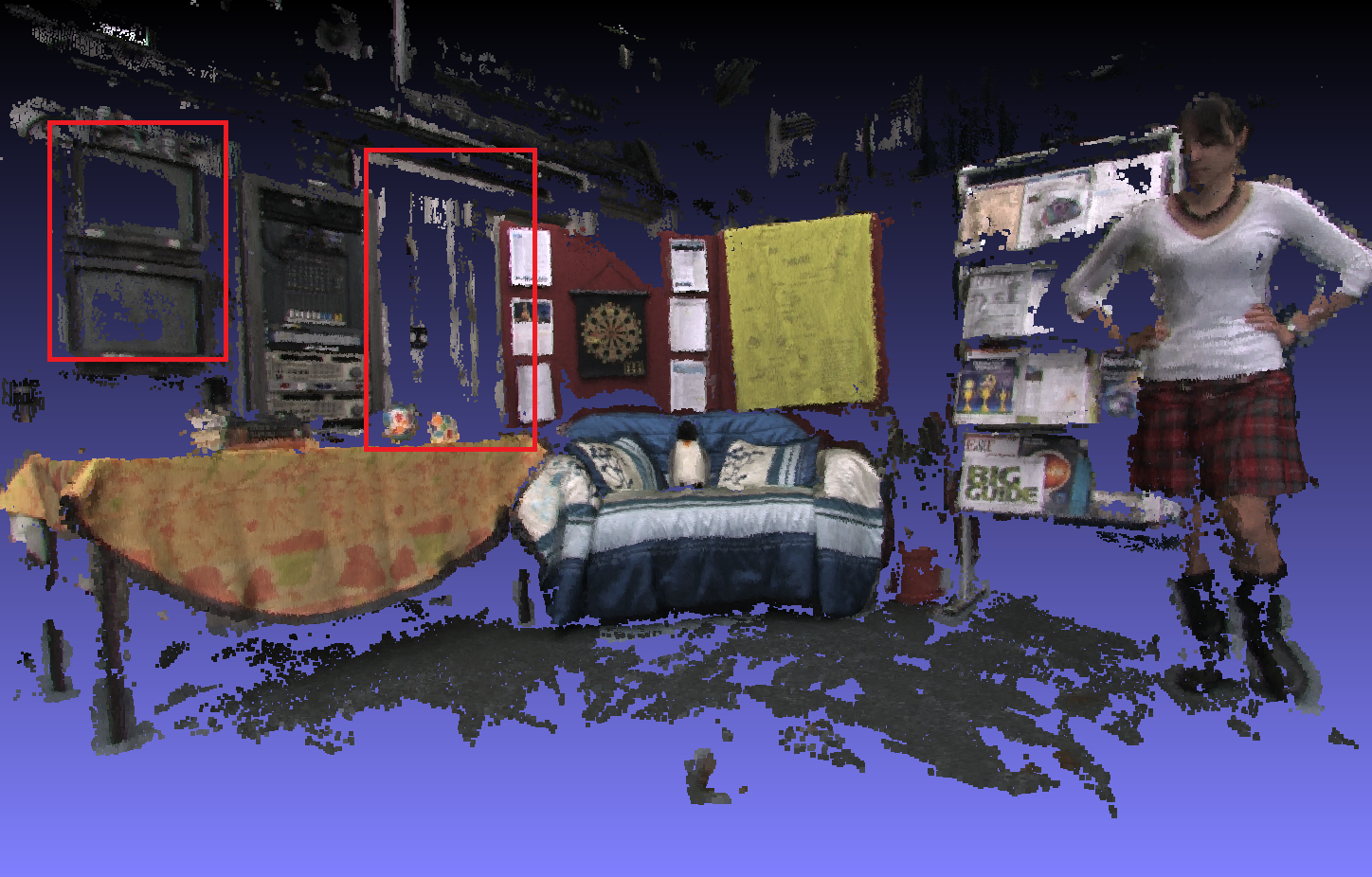}
        \label{fig:c}
    }
    \caption{Qualitative analysis of openMVG/openMVS pipeline with varying camera resolution and selection}     
    \label{nakedEyeComparison}
    \vspace*{-10pt}
\end{figure*}

%\vspace{-5pt}
\subsection{Problem Evidence Analysis} %{\color{red}(What is evidence?)}}
\label{section:problem}
Here, we use qualitative and quantitative experimental results to demonstrate the effects of frame resolution and the number of cameras on 3D reconstruction pipeline latency and quality in order to motivate the need for intelligent application- and system-side optimizations.
%Here we show experimental evidences through qualitative and quantitative benchmarking that simple fixes like frame resolution reduction and reducing the number of cameras for reconstruction drastically degrade the reconstruction quality and without significantly bringing down the reconstruction latency.
%{\color{blue}
%\textbf{Datasets and testbed used:}
We use publicly available Dance1~\cite{MustafaICCV15} dataset (7 synchronous video sequences from static cameras) %This original dataset contains concurrent images captured from 7 static cameras and 1 moving camera. For this work, we discard the data from the moving camera as it beyond the scope of this work. The data in each static camera contains 251 synchronized frames that is used for 3D reconstruction. 
%The recontruction experiments are performed 
on a Dell desktop with Intel i7 @2.9GHz processor with 16GB RAM %, and NVIDIA GeForce RTX 2060 
for the benchmarking experiments. The desktop used to some extent mimic the CPU capacity of a low-cost edge device. 
%as the front-end edge node. More details on the testbed will be explained later in Section~\ref{Sec:experiments}.
%{\color{red}Mention the Odzemok dataset as well.}

%Compute features &  5.49 & 5.46 & 5.47 & 5.45\\
%\hline
%Compute matches &  2.58 &  2.07 & 2.04 & 2.05\\
%\hline
%Structure from Known Poses &  0.95 & 0.68 & 0.64 & 0.69\\
%\hline
%Export to openMVS &  1.74 & 1.49 & 1.50 & 1.49\\
%\hline
%Densify point cloud &  \textbf{41.80} & \textbf{36.01} & \textbf{33.23} & %\textbf{35.95}    \\
%\hline
%Reconstruct the mesh &  13.30 & 12.02 & 10.14 & 12.22 \\
%\hline
%Refine the mesh &  20.56 &  17.45 & 15.04 & 17.88\\
%\hline
%Texture the mesh &  10.12 & 9.86 & 9.35 & 9.96\\
%\hline
%Total & 96.54 & 82.14 & 77.44 & 85.72\\
%\hline\hline
%\begin{tabular}{@{}c@{}}\textbf{Number of Vertices} \\(sparse/dense)\end{tabular} &  4159/11047 & 3585/9802 & 2847/8882 &  3554/10092\\
%%\hline
%%\textbf{F-score} & - & 0.98/0.81 &  0.95/0.75 &  0.983/0.85\\
%\hline
%\end{tabular}
%\end{table*}

\textbf{Quantitative  results:} Here we investigate the effects of simple and `quick fix' optimization strategies on processing latency. Table~\ref{quality} demonstrates how openMVG/openMVS pipeline latency can be reduced by varying camera resolution and number of cameras for reconstruction. The table compares the latency of  ideal configuration %(also referred as {\em golden results}) 
that includes all camera data with original resolution (scale = 1) against resolution compromised reconstructions. 
%The results show that openMVS step for dense reconstruction dominates the pipeline processing time i.e., the `Densify point cloud' takes up more than $90\%$ of total processing time. 
We observe that both the processing latency and the number of reconstructed vertices decrease as resolution decreases. This is because with lower resolutions, the images lose part of pixels during resizing.  % This causes the number of sparse and dense 3D vertices to significantly reduce as well as the processing time. 
Table~\ref{quality} also shows the impact of number of cameras and selection of cameras for reconstruction. In particular, we show the processing time comparison when we 
randomly remove camera $\#0$ and camera $\#2$ without altering the original resolution. The results demonstrate that the degree of such effect varies from camera to camera based on their orientation and coverage. For example, the effect of removing camera $\#2$ is greater than removing camera $\#0$ in terms of number of reconstructed vertices.

\textbf{Qualitative results:}
The qualitative analysis shows that the aforementioned latency minimizing `quick fixes' can adversely impact the quality of reconstructed 3D scene. 
%Fig.~\ref{nakedEyeComparison} shows the reconstructed scene quality with Dance1~\cite{MustafaICCV15} dataset. 
Fig.~\ref{fig:a} shows the golden results i.e., reconstruction with original resolution and all camera images. Whereas Fig.~\ref{fig:b} also uses all images but resizes images to scale $0.6$ (i.e., $60$\% of the original resolution). Fig.~\ref{fig:c} instead shows the reconstructed scene with original resolution images from all cameras but without images from camera \#2. Visually, we can see that with lower resolution, the reconstructed scene is not as smooth and delicate as the golden result (highlighted in Fig.~\ref{fig:b}). Similarly when using fewer images  (i.e., without images from camera \#2), the reconstructed scene is not as complete as the golden result (highlighted in Fig.~\ref{fig:c}). 
{\em The quantitative and qualitative results together motivate the need for intelligent optimizations towards meaningful reduction in reconstruction latency without significantly degrading quality.}

%}
%\begin{figure}[htb]
%    \centering
%    \includegraphics[width=0.45\textwidth]{img/opt_result_4.png}
%    \caption{The relationship between the number of vertices and pipeline processing time}
%    \label{fig:time_point}
%\end{figure}

\section{Optimization Methodology}
\label{Sec:optimization}
This section will discuss the overall optimization problem from two angles. First, we propose application-side optimizations involving background subtraction to reduce the processing time by removing repeated background information. Next, based on the application-side optimization outcome, we formulate a system optimization problem to address the trade-off between reconstruction quality and latency. This helps to find the optimal image resolution and the selection of cameras.
%This section discusses our approach to optimize latency that we attempted from application and system point of views. 

\subsection{Application-side Optimizations}
\label{sec:app_opt}
\begin{figure}[t]
    \centering
    \includegraphics[width=.4\textwidth]{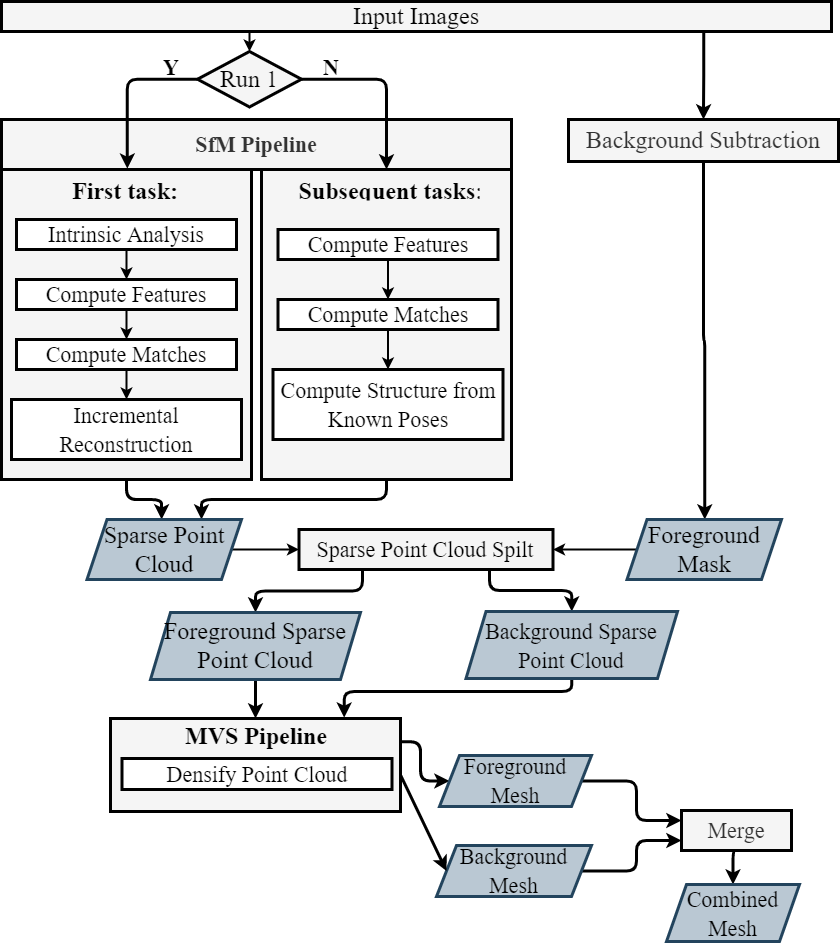}
    \caption{An overview of application implementation that runs parallel MVS pipelines for the foreground and background 3D reconstruction. }
    \label{fig:section4-1}
   % \vspace*{-10pt}
\end{figure}

In this paper, we propose a parallelized pipeline on openMVG/openMVS platforms by segmenting a scene into foreground (dynamic part) and background sections that can be later exploited by system-side optimizations through the collaborative MEC environment.
%The application side optimization is performed by segmenting a scene into foreground (dynamic part) and background sections that allowed us to reduce the effort for repetitive 3D reconstruction of the static part of the scene. Novel 3D reconstruction pipelines on top of openMVG+openMVS platforms are proposed for high frequency foreground reconstruction enabling rapid updates of the 3D map of a dynamic scene. 
This is achieved by the sub-tasks described below and illustrated as flowchart in Fig.~\ref{fig:section4-1}.   

%Since videos are captured by static cameras, the background of the scene remains unchanged. Therefore, it is unnecessary to reconstruct the background information repeatedly. We propose a novel reconstruction pipeline that generates a background mesh on its first task. The subsequent tasks only generate the foreground meshes and then merge the foreground meshes with the background mesh created by the first task.
%\iffalse
%\fi

\textbf{Background subtraction: }
%This is a technique to learn the static background model of a scene that can be subtracted from the current frame to identify the moving or dynamic regions. %Typically, the background model of a scene is formed by averaging an initial set of frames of a video. This would allow each pixel of the background model to learn the static parts of the scene because statistically, they reappear in the same places of the image frames which is unlike the dynamic parts.
We use openCV Background Subtractor to learn and subtract the static background model of a scene to identify the moving or dynamic regions. 
%The background-subtracted information is a binary image with white pixel regions suggesting dynamic regions and black pixels indicating the static regions. 
Since a scene can contain multiple moving objects, we used $k$-means clustering \cite{Lloyd1982} with optimum cluster numbers to group the dynamic regions into clusters. 
Each of the clusters suggests a moving object. 
This helps us to create separate foreground masks (contiguous dynamic regions in the background subtracted binary image) when moving objects are spatially apart, thus reducing the foreground area to be 3D reconstructed. 

\textbf{Sparse point-cloud split:}  %{\color{red}
Since openMVS requires a lot of computation to generate the dense point cloud~\cite{openmvs2020}, our optimized pipeline uses foreground masks (obtained from background subtraction) to group the sparse 3D points (obtained from openMVG) into the foreground and background sets. This separation enables running openMVS pipeline separately and concurrently for foreground and background 3D reconstruction. 
%} %In later sections, we discuss how high frequency foreground reconstruction and need basis background generation help us to significantly reduce the latency while maintaining high quality reconstruction results. {\color{red}Did not understand the last sentence.} \textcolor{cyan}{Can we delete the last sentence? It said that we would discuss [frequently reconstruct foreground] + [basis background reconstruction] significantly reduce the latency but maintaining the high quality.}

%Instead of separating the background and the foreground sections on images, we perform such separation  on the sparse point cloud obtained from openMVG. openMVS pipeline performs computational intensive dense point cloud generations, surface mesh construction, and adding color and texture information. Our optimized pipeline intelligently uses foreground masks to group the sparse 3D points into the foreground and background sets. This enables running openMVS pipeline separately for foreground and background 3D reconstruction.

\textbf{Merge foreground and background:} 
%\textcolor{cyan}{
Once the background and foreground 3D reconstruction results of consistent scale, translation, and rotation are stored in .ply 
files with the basic information of the 3D model such as, the number of 3D points and their locations and the color information of each point,
%, and which three points come up a face, etc.  
we generate the full 3D map of the scene by combining the information stored in the .ply files.

\vspace{-5pt}

\subsection{System-side Optimizations}
\label{sec:sys_opt}
%\vspace{-10pt}
% The system side optimization includes optimal configuration selection, camera selection algorithm, background update strategy, and overall implementation on an inter-edge collaboration architecture. 
%As demonstrated in Section~\ref{section:problem}, adjustments such as scaling down the resolution and reducing the number of cameras can further cut down the processing time. However, such adjustments will inevitably lead to a certain quality degradation. 
For system optimizations, we propose an {\em online} optimization framework to address the trade-off between quality and processing time by choosing optimal reconstruction configurations. This framework exploits the collaborative MEC environment to achieve task level parallelism.

\textbf{Problem formulation:}
% While background subtraction moderately reduces the pipeline processing time, there is still a lot to be done from the system point of view to improve latency reduction even further. As demonstrated earlier, we can further reduce the processing time by reducing the camera resolution and number of cameras. However, such adjustments will inevitably lead to a certain quality degradation. Here we propose an optimization framework to address the trade-off between quality and processing time by choosing optimal reconstruction configuration, i.e., camera resolution and number of cameras. Specifically, the framework aims to find this optimal configuration during the run-time, i.e., {\em online}. 
We define $\mathcal{N} = \{1,2,3,..N\}$ to be the set of cameras (i.e., edge devices for video capture), where $N \geq 3$. We create a binary camera-selection indicator $o_n, \forall n \in \mathcal{N}$ i.e., $o_n = 0$ indicates that the image from camera-$n$ will be ignored by the pipeline; otherwise $o_n = 1$. We also denote $\Delta_t$ as the user's (i.e., reconstruction task's) deadline requirement. Based on the problem evidence analyses, we know that both the quality $Q$ and the processing latency/time $T$ are functions of camera/image resolution $r\in[r_{min},r_{max}]$ and camera subset $\mathcal{N}' \subseteq \mathcal{N}$, i.e., $Q(r, \mathcal{N}')$ and $T(r, \mathcal{N}')$. With such settings, our optimization problem can be stated as:
\begin{equation}
 \begin{aligned} 
  \max_{r, \mathbf{o}} & \; Q(r, \mathcal{N}')
  \\
  \textbf{s.t.} \:\:
  \textbf{C1: } & r\in [r_{min},r_{max}]
  \\
  \textbf{C2: } & o_n \in \{0,1\},\;\forall n \in [1,N]
  \\
  \textbf{C3: } & \sum\limits_{n=1}^{N} o_n \geq 3,\;\forall n \in [1,N]
   \\
  \textbf{C4: } & T(r,\mathcal{N}') \leq \Delta_t
\end{aligned}
\tag{\textbf{P1}}  \label{Eq:p1}
\end{equation} 
where constraints \textbf{C1} and \textbf{C2} restrict the selection of camera resolution and number respectively, \textbf{C3} specifies that the  number of cameras should be at least 3 (required by openMVG~\cite{moulon2016openmvg}), and \textbf{C4} specifies that the processing time needs to satisfy the user's deadline requirement $\Delta_t$. %, i.e., the time interval between two consecutive tasks.
 Problem \textbf{P1} is non-trivial to solve as: 1) Firstly, it is time-consuming to create accurate mathematical models (such as by taking massive measurements~\cite{wang2020joint}) for quality $Q(r, \mathcal{N}')$ and processing time $T(r, \mathcal{N}')$ %2) Secondly, the application itself is data-content aware, i.e., the value of $Q(r, \mathcal{N}')$ and $T(r,\mathcal{N}')$ may change all the time even with the same configurations. This fluctuation affects the effectiveness of the optimization; 
 and 
2) Secondly, the camera indicators are binary variables and the number of camera subset is very large (i.e., $\mathcal{O}(2^{N}$)). Therefore, computing all the solutions in this massive configuration space at run-time will be counterproductive.

\textbf{Camera selection algorithm:} 
In order to reduce the complexity of \textbf{\eqref{Eq:p1}}, we propose a camera selection algorithm to select the $N'$ most useful cameras. However, the camera selection scheme with 3D reconstruction is quite different from the typical maximal coverage problems~\cite{wei2015continuous} as in 3D reconstruction, the reconstructed points must be covered by at least two cameras. In this algorithm, the 3D points and their corresponding cameras are first extracted from the SfM pipeline. With these, our objective is to choose the cameras that will result in the maximum number of 3D points that are covered by at least two cameras. The most trivial way to do this is to implement a brute force technique; however, that will cause the running time to grow exponentially with the increase in number of cameras.
Therefore, we develop the following optimization problem:
\begin{equation}
 \begin{aligned} 
  \max_{} & \sum_{k=1}^{\mathbb{P}} p_{k}
  \\
  \textbf{s.t.} \:\:
  \textbf{C1: } & \sum_{n=1}^{N} o_{n} = N'
  \\
  \textbf{C2: } & M_{kn} = c_{n} A_{kn},\; \forall  n \in [1,N],\; \forall k \in [1,\mathbb{P}]
  \\
  \textbf{C3: } & \sum_{n=1}^{N} M_{kn} \ge 2p_{k} ,\; \forall k \in [1,\mathbb{P}]
\end{aligned}
\tag{\textbf{P2}}  %just a placeholder, please change tag to make more sense
\label{opt_dec}
\end{equation} 
In \textbf{\eqref{opt_dec}}, we assume that there are a total of $\mathbb{P}$ 3D points and $N$ cameras and out of them $N'$ cameras are chosen. $p_{k}$ is a binary variable which is 1 if 3D point-$k$ is covered by at least two \textit{chosen} cameras and $0$ otherwise. Thus the objective function is to maximize the points that are covered by at least two cameras. %As mentioned before $o_{n}$ is a binary variable which is 1 if camera-$n$ is chosen and 0 otherwise; thus, 
The constraint \textbf{C1} states that a total of $N'$ cameras are chosen.
Also, we assume that $A_{kn}$ is a \textit{known} binary variable that is 1 if point-$k$ is covered by camera-$n$ and 0 otherwise; and $M_{kn}$ is a binary variable that is 1 if (a) point-$k$ is covered by camera-$n$, and (b) camera-$n$ is chosen, which is ensured by constraint \textbf{C2}. 
Constraint \textbf{C3} ensures that point-$k$ is covered by at least two chosen cameras. We use Gurobi solver%~\cite{gurobi} 
to solve \textbf{\eqref{opt_dec}}. 
As the number of 3D points is pretty large in a complex image, we first choose a small fraction of \textit{key} points using $k$-means algorithm, and apply \textbf{\eqref{opt_dec}} with these key-point subset to choose the most useful cameras.

%The overhead for initializing the solver was very high when using the entire set of 3D points. In order to diminish this, we used mini batch kmeans to effectively reduce the number of 3D points to a key-point subset. 
%\par Despite the introduction of the better optimized algorithm, the brute force algorithm still outperforms it with smaller amounts of cameras. So it is best to use a hybrid algorithm, where it uses the brute force algorithm when the number of cameras is less than 12 and the optimized algorithm \ref{opt_dec} otherwise.

\begin{algorithm}[htb]
\footnotesize
%\begin{footnotesize}
\caption{Online Bi-section search algorithm}
\label{Algo:1}
%\begin{algorithmic}
\textbf{Initialization: } Set $N'=N$, $r_{min}=0.3$, $r_{max}=1.0$, $r^{*}=1.0$;\\
Run the first task with $(r^*, \mathcal{N})$,  solve \textbf{P2} and initial $\pi(N'): N' \rightarrow \mathcal{N'}$;\\
$r=r_{min}$, $solutions=[]$;\\
%\textit{baseline\_check} = True, 
$optimization$ = True;\\
\For{each upcoming task $i \in \mathcal{I}$}
{   
   Observe $T(r, \pi(N^\prime))$, $Q(r, \pi(N^\prime))$ \\
   \If{\textbf{not} $optimization$}
    {   
        %Observe $T(r^*, \pi({N^\prime}^*))$, %$Q(r^*, \pi({N^\prime}^*))$ %\tcp*{solve % \textbf{P3}}
       % \\
        do \texttt{minor-adjustment}  \tcp*{see section IV-B} 
        %\\
        \textbf{continue}
    }
   % }\Else{
%         Observe $T(r, \pi(N^\prime))$, $Q(r, %\pi(N^\prime))$ %\tcp*{solve \textbf{P3}}
%         \\
%    }
   %\textit{reduce\_camera\_number} = False;\\
   \uIf{$T(r, \pi(N^\prime)) \le \Delta_t$}
   {
       $r_{min}=r$; \:\:   $r^*=r$; \:\:  \\ %\textit{baseline\_check} = False; 
       $solutions$ $\leftarrow$ [$Q(r, \pi(N^\prime))$] \tcp*{stores the current optimal configurations}
   }\Else{
        $r_{max} = r$;
   }
   \uIf{$r_{max}-r_{min}\leq \tau$}
     {
         $r_{min}=r^*$; \:\:  $r_{max}=1.0$;\:\:
         $r=r_{min}$;\:\:       $N^\prime = N^\prime-1$\tcp*{another level of search} 
         \If{$N^\prime < 3$ }
        {  
           $r^*, {N^\prime}^* = \argmax(Q(r, \pi(N^\prime)))$ in $solutions$     
          $optimization$ = False;
        }
     }\Else{
        $r = (r_{max}+r_{min})/2$ 
     }
}
\end{algorithm}

\textbf{Transformation of P1 and solution:}
After solving \textbf{\eqref{opt_dec}}, we create a mapping function $\pi(N'): N' \rightarrow \mathcal{N'}$ that specifies the unique mapping between the number of cameras and the selection of camera subset, i.e., given the number of cameras $N'$, the selection of $\mathcal{N'}$ is fixed. This function greatly reduces the domain size of \textbf{\eqref{Eq:p1}}. %In addition, since $Q(r, \mathcal{N}')$ and $T(r,\mathcal{N}')$ have the data-varying feature, we will focus on their expectations. 
Therefore, \textbf{\eqref{Eq:p1}} transforms to: 
\begin{equation}
 \begin{aligned} 
  \max_{r, N'} & \; Q(r,\pi(N'))
  \\
  \textbf{s.t.}  \:\:
  \textbf{C1: } & r\in [r_{min},r_{max}]
  \\
  \textbf{C2: } & N' \in [3,N]
   \\
  \textbf{C3: } & T(r,\pi(N')) \leq \Delta_t
\end{aligned}
\tag{\textbf{P3}}  \label{Eq:p2}
\end{equation} 
In \textbf{\eqref{Eq:p2}}, $Q(r,\pi(N'))$ and $T(r,\pi(N'))$ are still unknown to the MEC system. However, by fixing one of the two parameters (e.g., $Q(r | \pi(N'))$, $T(r | \pi(N'))$ and $Q( \pi(N') | r)$, $T(\pi(N')|r)$), we can easily get that the quality and the processing time are monotonic functions of both $r$ and $N'$. Therefore, the problem can be effectively solved by a two-dimensional Bi-section algorithm. The algorithm is driven by online observations that is described in Algorithm~\ref{Algo:1}. %{\color{red}In algorithm 1, please change section 4.2.3}{\color{blue}changed}

In Algorithm~\ref{Algo:1}, we perform multiple Bi-section searches; the goal of each Bi-section is to find the optimal resolution $r^*$ for a given camera subset $\pi(N')$. We assume the minimum resolution $r_{min}$ to be 0.3. The algorithm starts with a default configuration $(r=r_{min}, \pi(N'=N))$ (lines 3-4). For each upcoming task, the MEC system first runs the pipeline and observes the processing time $T(r, \pi(N^\prime))$ and quality $Q(r, \pi(N^\prime))$ (line 6). In lines 10-21, the algorithm compares the current processing time to the deadline $\Delta_t$ and performs Bi-section search on the resolution for the current camera subset $\pi(N^\prime)$. In the meantime, it stores the current optimal configurations $(r^*, \pi(N^\prime))$ and the corresponding quality into $solutions$ vector (line 12). In lines 15-16, when the search on current camera subset be finished, the algorithm resets the $r_{min}$ to the current optimal resolution $r^*$. Next it goes into the next Bi-section search on camera subset $\pi(N'=N'-1)$. This process terminates when $N'<3$ or $r_{max}-r_{min}\leq \tau$, where $\tau$ is a constant which is assumed to be 0.01.

We also add the following functions to further reduce the time complexity during the configuration exploration.
\begin{enumerate}[leftmargin=*,itemsep=0pt]
    \item \textit{Baseline\_check:} If $T( r_{min}|\pi(N'_i)) > \Delta_t$, stop Bi-section search on $N'_i$ and go to $N'_i-1$. This means that the processing of the lowest resolution has exceeded the deadline and there is no need to further explore $N'_i$, i.e., $T( r' |\pi(N'_i)) > T( r_{min}|\pi(N'_i))$, $\forall r'> r_{min}$.
    \item Given $N^{\prime}_i < N^{\prime}_j$, if there exists a feasible solution ($r^*_j, N^{\prime}_j$) and $Q( r_{max}|\pi(N'_i)) \leq Q( r_j^* | \pi(N'_j))$, terminate optimization. The reason being $Q( r_{max}|\pi(N'_k)) \leq Q( r_j^* | \pi(N'_j))$, $\forall N'_k < N'_j$.
    \item Given $T( r|\pi(N'_i)) > \Delta_t$, if there exists a feasible solution ($r^*_j, N^{\prime}_j$) and $Q( r|\pi(N'_i)) \leq Q( r_j^* | \pi(N'_j))$, stop Bi-section search on $N'_i$ and go to $N'_i-1$. The reason being $Q( r'|\pi(N'_i)) \leq Q( r_j^* | \pi(N'_j))$, $\forall r' < r$.
\end{enumerate}

%{\color{blue}
\textbf{Minor-adjustment procedure:} We add a \textit{minor-adjustment procedure} (line 8 in Algorithm~\ref{Algo:1}) to solve the problem of processing time fluctuation between consecutive tasks and facilitate the average processing time to converge to $\Delta_t$.  At a given timestamp $I'\leq I$, the adjustment rule follows:
\begin{equation}
     r^* \text{=}
     \begin{cases}
     r^* + \tau &  {1\over I'}\sum\limits_{i=0}^{I'}T_i(r^*) < \Delta_t \;\&\; T_i(r^*) < \Delta_t
     \\
     r^* - \tau & {1\over I'}\sum\limits_{i=0}^{I'}T_i(r^*) > \Delta_t \;\&\; T_i(r^*) > \Delta_t
     \\
     r^* & \text{otherwise}
     \end{cases}
 \end{equation}
Based on the difference between current average processing time and deadline, we slightly decrease or increase $r^*$ by $\tau$ (e.g., 0.02).

\begin{figure}[t]
    \centering
    \includegraphics[width=.48\textwidth]{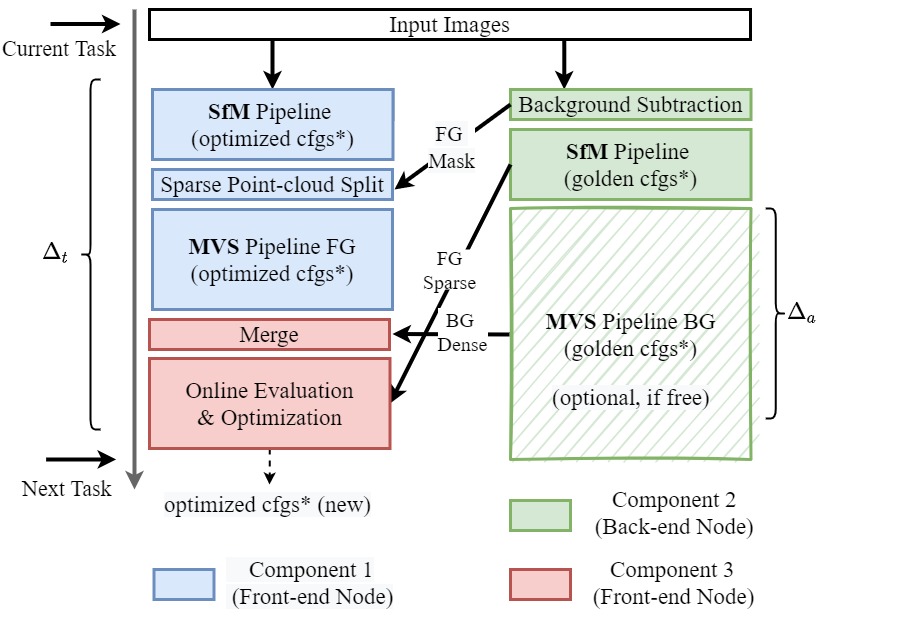}
    \caption{An overview of the collaborative MEC environment running the optimized pipeline}
    \label{fig:new-pipeline}
    %\vspace*{-12pt}
\end{figure}

\textbf{Background update strategy:}
\label{sec:bg_update}
As explained in Section~\ref{sec:app_opt}, the application-side optimization through background subtraction updates the foreground frequently and therefore is reconstructed for each task. However, the background reconstruction is significantly more time consuming and computationally intensive. Thus, a continuous background reconstruction in parallel to foreground is only going to lengthen the overall processing time. Contrarily, the polar opposite method of performing the background reconstruction only once during the entire reconstruction lifecycle can benefit processing latency; however, might significantly degrade the reconstruction quality.
%Since background reconstruction for each task is significantly more time consuming and computationally intensive. Besides, the background areas change less frequently compared to the areas in foreground. Therefore, we allow the system to reconstruct the foreground for every task while reconstructing the background at a low frequency. 
To address this issue, we let the front-end node of the collaborative MEC environment to continuously reconstruct the foreground and the back-end edge node to opportunistically perform background reconstruction only during its idle time (denoted by $\Delta_a$ as shown in Fig~\ref{fig:new-pipeline}). The newly reconstructed background result will be actively pushed to the front-end node.

\textbf{Overall system implementation:} 
The resulting optimized framework implementation with all application-side and system-side optimization steps on the collaborative MEC environment is illustrated in Fig.~\ref{fig:new-pipeline}. %The inter-edge collaboration architecture uses the two collaborative edge nodes (front-end and back-end) as described in the system model (in Section~\ref{Sec:system-model}). 
First, the front-end node performs ``SfM Pipeline''  with optimized configurations, while the back-end node simultaneously executes ``Background Subtraction'' and sends the foreground masks to the front-end node. Once it gets the foreground masks, the front-end node starts  ``Sparse Point-cloud Split'' and ``MVS pipeline FG'' to create the dense foreground point-cloud, and then merges it with the existing background point-cloud. During the same time, the back-end node performs the ``SfM pipeline'' with golden configurations, and the result is sent to the front-end node for ``Online Evaluation $\&$ Optimization''  (we discuss online evaluation in detail in Section~\ref{subsec:eval_method}). Beyond that, the back-end node uses spare time to reconstruct the background until the next task arrives as explained in Section~\ref{sec:app_opt}. The processing latency of the proposed  pipeline depends on the length of its `critical path', which is the time elapsed between component 1 and component 3.

\begin{table*}[t]
\caption{Foreground F-score comparison between {\em online} and {\em offline} evaluations}
%\vspace*{-6pt}
\label{opt_evl_proof}
\centering
\begin{footnotesize}
\begin{tabular}{|c||c|c|c|c|c|c|}
\hline
\bfseries F-score & 
\bfseries 
\begin{tabular}{@{}c@{}}Scale=0.95 \end{tabular}
& \bfseries 
\begin{tabular}{@{}c@{}}Scale=0.90\end{tabular}
& \bfseries 
\begin{tabular}{@{}c@{}}Scale=0.85 \end{tabular}
& \bfseries 
\begin{tabular}{@{}c@{}}Scale=0.80 \end{tabular}
& \bfseries 
\begin{tabular}{@{}c@{}}Scale=0.75 \end{tabular}
& \bfseries 
\begin{tabular}{@{}c@{}}Scale=0.70 \end{tabular}
\\
\hline\hline
%Processing Time  & 38.70 & 32.24 & 27.63 & 24.64 %\\
%\hline
After openMVG ({\em online evaluation})  & 0.911 & 0.888 &  0.873 & 0.864 & 0.855 & 0.819 \\
\hline
After openMVS ({\em offline evaluation})  & 0.873 & 0.851 & 0.836 &  0.817 & 0.797 & 0.770 \\
\hline
\end{tabular}
\vspace*{-6pt}
\end{footnotesize}
\end{table*}

% 0.75 = [0.797, 0.784]
% 0.80 = [0.817, 0.820]

%{\color{sdr}
\section{System Evaluation and Results}
\label{Sec:experiments}
In this section, we evaluate the performance of the proposed optimized framework and validate our {\em online} quality evaluation method through experiments on a hardware edge testbed with real datasets. The hardware MEC testbed mostly mimics the system model from Fig.~\ref{fig:system_model} and consists of a front-end node (Dell desktop equipped with Intel i7-10700F @2.9GHz, 16GB RAM)  % and NVIDIA GeForce RTX 2060
and a back-end node (Dell desktop equipped with Intel i7-10700k @3.8GHz, 32GB RAM). These mimic low cost edge devices with no GPU capability. % and NVIDIA GeForce RTX 2060
The two nodes are connected via 10 Gbps Ethernet cable mimicking point to point connectivity between the nodes.  
%{\color{blue}
%Since this paper mainly focuses on optimizing computational performance and the network delay between the user site and edge site does not impact the overall performance {\color{red}(May be a bit strong statement)}, our testbed does not include the user site (from the system model) and we assume that the videos are pre-stored in the front-end node.}
%{\color{red}
The video datasets used for the evaluations are Dance1 and Odzemok~\cite{MustafaICCV15}. 
%} 
%WHY? NEEDS MORE DETAILS OF THE TESTBED. HOW AND WHY IS IT DIFFERENT FROM FIG. 3?}

 %The system prototype is described in Fig.~\ref{fig:hardware}. 

%\begin{figure}[htb]
%\vspace*{-12pt}
%    \centering
%    \includegraphics[width=.49\textwidth]{img/hardwa%re.png}
%    \caption{Hardware Description.}
%    \label{fig:hardware}
%    \vspace*{-6pt}
%\end{figure}

%We also validate our {\em online} quality evaluation method and the performance of the proposed optimization algorithm and background reconstruction strategy.  The dataset used for the evaluations are Dance1 and Odzemok from CVSSP-3D Project Sample Datasets~\cite{MustafaICCV15}. 

\subsection{Evaluation Method}
\label{subsec:eval_method}
For {\em offline} qualitative evaluation, i.e., to test the performance of our optimized pipeline (without the concern of evaluation latency), we use the metric proposed in~\cite{ruanobenchmark}. In this method, the quality is measured in terms of precision ($P$), recall ($R$), and F-score ($F$) - where $P$ measures how close a 3D point cloud is to the ground truth, $R$ measures the completeness of the reconstruction, and $F$ is a function of $P$ and $R$, i.e. ,
\begin{equation*}
    F = {2PR\over(P + R)}
\end{equation*}
In this paper, we select distance threshold $d$ = 0.01 mm, 0.02 mm
%{\color{red}(How are you getting mm \?} 
to determine whether a point from the reconstructed point cloud and a point from the ground truth are close enough. 
However, due to the lack of ground truth, we run the original openMVG/openMVS pipeline to reconstruct a 3D model with the best configuration (i.e., original resolution at scale=1 and with all cameras). This {\em golden result} (as explained in Section~\ref{Sec:system-model}) is our best estimate of the ground truth and thus is used to calculate the F-score of the reconstructed outcome of our optimized pipeline. %However, due to the inherent randomized elements within the openMVG and openMVS algorithms, the reconstructed scenes between runs might contain minor inconsistencies. Therefore, {\em golden results} between runs can also fluctuate. For example, in dataset Dance1, the F-score of the {\em golden results} from original pipeline typically has a mean of $0.877$ with standard deviation of $0.007$. This is calculated over 10 different runs where the ground truth for one particular run was the reconstructed outcome of the previous run. The same values for dataset Odzemok are $0.832$ and $0.016$. We will keep this aspect of fluctuating ground truth in mind when we later use them to evaluate the quality of our optimized pipeline.   

%, denoted as $G$ (comparable to the ground truth), which means this golden result is the standard when we tend to evaluate the later 3D models, the 3D model to be evaluated is denoted as $I$.
%\vspace{-0.3in}
\subsection{Validity of Online/Run-time Quality Evaluation}
One of the trickiest aspects of our optimized pipeline is to measure the quality $Q(r,\pi(N'))$ of 3D reconstruction {\em online}, i.e., in runtime and use this outcome to ascertain optimal configurations for next task. In order to achieve this, we perform the evaluation on the foreground sparse point cloud obtained from openMVG's SfM pipeline instead of the mesh from openMVS (as shown in Fig.~\ref{fig:new-pipeline}) due to the fact that: i) the former is the foundation of the latter; ii) openMVS steps often take more time; and iii) the foreground is more important and always changing. Whereas, for the {\em offline} qualitative evaluations explained in Section~\ref{subsec:eval_method}, we take the more traditional approach of comparing the foreground mesh from openMVS against the foreground from {\em golden result}. In order to establish the validity of the proposed {\em online} evaluation by establishing positive correlation between the {\em online} and {\em offline} evaluations, we compare the F-scores values of foreground after openMVG's SfM against foreground after openMVS in Table~\ref{opt_evl_proof}. The results show that for both cases F-scores decrease as resolution decreases. Furthermore, their decreasing trends are also comparable. Therefore, we believe that our proposed {\em online} evaluation is a valid methodology that successfully reflects the quality of the 3D reconstruction of our optimized pipeline.
%The dataset used for this experiment is Dance1. 

%To evaluate the reconstruction quality of each task simultaneously while the pipeline is running (online evaluation), we perform the evaluation procedure on the foreground sparse point cloud obtained from openMVG instead of the mesh got from openMVS, because 1) the former is the foundation of the latter; 2) openMVS steps often take much more time; 3) the foreground is more important and always changing. Whereas, for the off-line evaluation (to test the performance of our pipeline without concerning the completion time), we compare the resulting foreground mesh of openMVS with the foreground golden result. By our observation, the results of the online and off-line evaluation are positively correlated under the same configuration. 

%As shown in Table~\ref{opt_evl_proof}, the Sparse F-score (openMVG result of the foreground), and Dense F-score (openMVS result of the foreground) are decreasing as the resolution decreases. Furthermore, their decreasing trends are pretty similar. Therefore, we believe that the online evaluation can reflect the quality of the 3D reconstruction of our pipeline.

\begin{figure}[t]
    \centering
    \includegraphics[width=.5\textwidth]{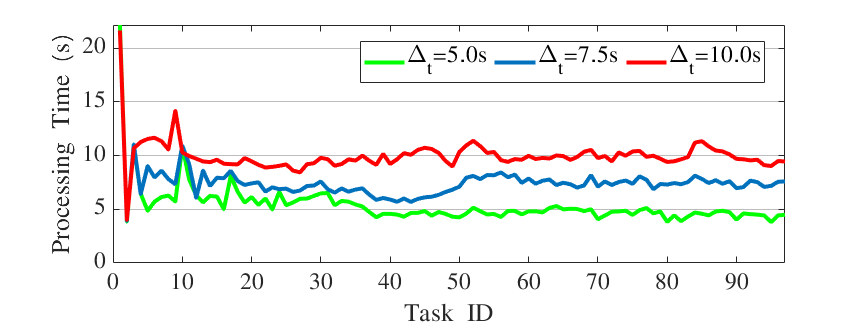}
    \caption{Task processing time deadline satisfaction}
    \label{fig:opt1_time}
   % \vspace*{-6pt}
\end{figure}

\begin{figure}[t]
    \centering
    \includegraphics[width=.5\textwidth]{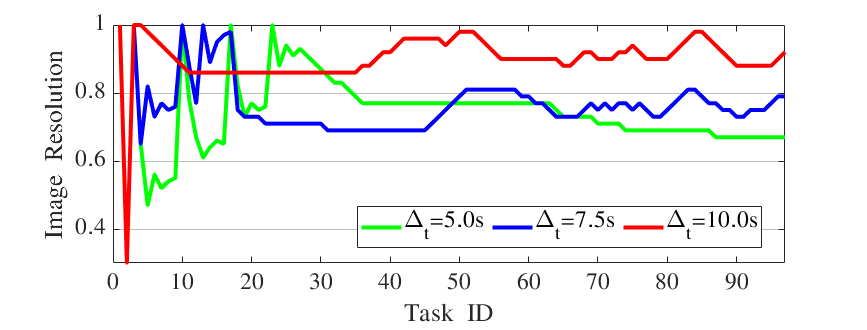}
    \caption{Resolution adaptation.}
    \label{fig:opt1_R}
    %\vspace*{-6pt}
\end{figure}

\begin{figure}[t]
    \centering
    \includegraphics[width=.5\textwidth]{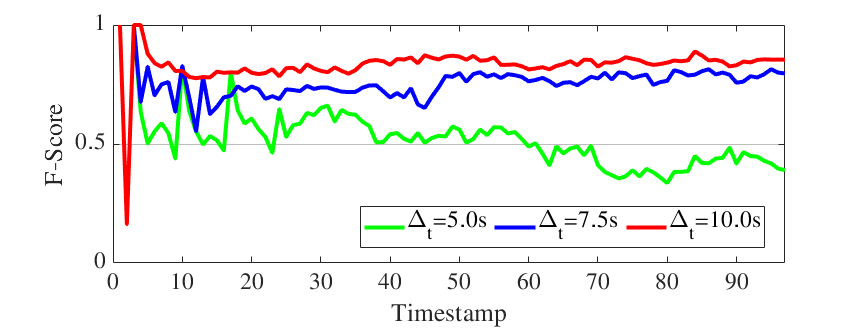}
    \caption{F-score (foreground only).}
    \label{fig:opt1_F}
    %\vspace*{-6pt}
\end{figure}

\subsection{Optimization Evaluation}
%{\color{blue}
Given a processing deadline $\Delta_t$, here we evaluate the outcome of the online optimization and background update strategy in terms of average task processing time and reconstruction quality $Q(r,\pi(N'))$ (using F-score). Taking into account that the processing time $T(r,\pi(N'))$ of the pipeline can vary according to the content of the input images, we accept a small margin of error during the configuration searching steps (based on experimental results, the margin is set to 1s). Specifically,  a configuration is considered as acceptable if $0 < |\Delta_t - T(r,\pi(N'))| \leq 1$. In this evaluation, the deadlines $\Delta_t$ of test cases are 5.0s, 7.5s and 10.0s, respectively.

Fig.~\ref{fig:opt1_time} shows the  processing time of individual tasks at different timestamps. Driven by the minor-adjustment procedure that we proposed in subsection~\ref{sec:sys_opt}, we notice that the processing time gradually converges to the pre-defined deadline $\Delta_t$ after the optimal configuration is found. The results of optimal configuration search (\textbf{Alg. 1}) and subsequent resolution adaptation (minor-adjustment procedure) are shown in Fig.~\ref{fig:opt1_R}. As the system takes more configuration search steps to find the optimal number of cameras for tasks with shorter deadline, the figure demonstrates that the image resolution of tasks with longer deadline (e.g., $\Delta_t=7.5s$ and $\Delta_t=10.0s$) converges faster than tasks with shorter deadline (e.g., $\Delta_t=5.0s$).
 %{\color{green}(Not clear)}  
 Fig.~\ref{fig:opt1_F} shows the quality distribution $Q(r,\pi(N'))$ (in terms of F-scores) of the foreground. After some initial randomness while the online optimization algorithm is still running, the F-score values of foreground sparse 3D point-cloud start to converge once the minor adjustment procedure starts to work. Such convergence behavior of F-score under different task deadlines is similar to resolution adaptation (as shown in Fig.~\ref{fig:opt1_R}). The overall statistical outcome in terms of chosen configuration along with average processing time, average F-score,  background update and number of search steps are presented in Table~\ref{opt_compare}.

\begin{table}[t]
\caption{Online optimization with different deadlines - Dance1}
%\vspace*{-6pt}
\label{opt_compare}
\centering
\begin{footnotesize}
\begin{tabular}{|c||c|c|c|}
\hline
\bfseries Info & \bfseries 
\begin{tabular}{@{}c@{}}Deadline \\($\Delta_t=5s$)\end{tabular}
& \bfseries 
\begin{tabular}{@{}c@{}}Deadline  \\($\Delta_t=7.5s$)\end{tabular}
& \bfseries 
\begin{tabular}{@{}c@{}}Deadline  \\($\Delta_t=10s$)\end{tabular}
\\
\hline\hline
Avg. Resolution Scale & 0.74 & 0.76 & 0.90  \\
\hline
$\#$ Cameras & 4 & 7 & 7 \\
\hline
Avg. Processing Time & 5.28 s & 7.49 s & 9.98 s  \\
\hline
Avg. F-score of FG & 0.51 & 0.74 & 0.83 \\
\hline
\begin{tabular}{@{}c@{}}$\#$ Reconstructed BG\\Dense Point Clouds\end{tabular}    & 22 & 36 & 54 \\
\hline
\begin{tabular}{@{}c@{}}$\#$ Configuration Search\\Steps\end{tabular}   & 27 & 12 & 1 \\
%\hline
%\begin{tabular}{@{}c@{}}Average task success rate\\ (Overall/Minor-Adjustment Area)\end{tabular} & 64\%/86\% & 69\%/81\% \\
\hline
\end{tabular}
%\vspace*{-6pt}
\end{footnotesize}
\end{table}

%minor-adjustment procedure

%The table also contains similar statistics for another experimental scenario with task processing deadline $\Delta_t =10 s$. We observe that for this scenario, the chosen optimal configuration is $(r^*=1.0, N^{\prime}=4)$ with mean processing time and F-scores are $10.82$ s and $0.69$ respectively.    
%the deadline $\Delta_t$ is set to $25 s$. The optimization takes 11 iterations. The optimal resolution is $r=0.7$ and number of cameras is $N^{\prime}=6$. From task $ID=12$ to task $ID=60$, the system terminates the optimization and uses minor-adjustment to reduce the amplitude of fluctuations in processing time. The average processing time and F-score of foreground sparse 3D point-cloud (from online evaluation) are $25.41s$ and $0.75$, respectively. The number of failed tasks is 11 but most of them are from task $ID=1$ to task $ID=11$. With $(r=0.7, N^{\prime}=6)$ and the minor-adjustment, only 3 tasks failed. When $\Delta_t =30 s$, the optimal solution is $(r^*=0.77, , N^{\prime}=7)$. These results as shown in Table~\ref{opt_compare} highlight the ability of our optimization in handling different user’s requirements.
%{\color{blue}

While evaluating the performance of the background update strategy discussed in Section~\ref{sec:sys_opt}, we observe that for datasets such as Dance1, the system only needs to create a background model once as the background scene remains unchanged. However, for real-world scenarios where background scene may change frequently, it is beneficial to use up-to-date background model. Upon monitoring the speed of background update for dataset Dance1 (with different processing deadlines), we observe that the back-end node can generate a new background point cloud around $20-25$s without interrupting the foreground reconstruction. Moreover, the background update frequency of tasks with larger deadline is much higher than tasks with shorter deadline. %{\color{red}
For example, the number pushed background dense point clouds are 22, 36 and 54 when the task deadline is set to $\Delta_t=5s$, $\Delta_t=7.5s$ and $\Delta_t=10s$, respectively.
%}
%For example, when $\Delta_t=10s$, the back-end node pushes 54 background dense point clouds to the front-end node. While the number of pushed point clouds reduced to 36 and 22 as the deadline is set to 7.5s and 5.0s, respectively. 
This can be explained by Fig.~\ref{fig:new-pipeline} i.e., larger deadline $\Delta_t$ leads to larger $\Delta_{\alpha}$. %{\color{red} 
In other words, while the front-end node is busy running foreground reconstruction with some high configurations,  the back-end edge node has more idle time to perform background reconstruction.
%}.
The above results indicate that the proposed online algorithm (\textbf{Algo.}~\ref{Algo:1}) and optimized pipeline (Fig.~\ref{fig:new-pipeline}) can effectively balance the trade-off between latency and quality.% based  on  the  availability of MEC resources.

\begin{table}[t]
\caption{Quality comparison between original and optimized pipeline}
%\vspace*{-6pt}
\label{golen_compare}
\centering
\begin{footnotesize}
\begin{tabular}{|c||c|c|}
\hline
%\bfseries 
{\bf F-score} & 
\textbf{Threshold} $d=0.01$
& \textbf{Threshold} $d=0.02$
\\
\hline\hline
\begin{tabular}{@{}c@{}}{\bf Optimal} \\ Dance1\end{tabular} & \begin{tabular}{@{}c@{}}avg F-score = 0.930 \\std = 0.026\end{tabular}  & \begin{tabular}{@{}c@{}}avg F-score = 0.961 \\(std = 0.019)\end{tabular} \\
\hline
\begin{tabular}{@{}c@{}}{\bf Optimal} \\ Odzemok\end{tabular} & \begin{tabular}{@{}c@{}}avg F-score = 0.946 \\(std = 0.012)\end{tabular}  & \begin{tabular}{@{}c@{}}avg F-score = 0.967 \\(std = 0.011)\end{tabular} \\
\hline
\end{tabular}
%\vspace*{-0.1in}
\end{footnotesize}
\end{table}

\subsection{Quality and Latency Evaluation}
Finally, we compare the quality and latency results of our optimized pipeline against the {\em golden results} of the original pipeline for both the datasets. For the quality comparisons, we run both pipelines with the same input images and compute F-score - when construct with original resolution and all camera data. The F-score for the optimized pipeline is calculated by assuming the {\em golden results} (i.e., from original pipeline) as ground truth.
Table~\ref{golen_compare} shows the average F-scores and standard deviation for each of the datasets.
Overall, we can conclude that reconstruction quality from our optimal pipeline are within $4- 7$\% of the peak quality (from Table~\ref{golen_compare}) which is negligible for successful operation of other AR/VR/MR applications that typically run after reconstruction pipeline.
However, when we compare the latency results of our optimized pipeline against original pipeline (as shown in Table~\ref{golen_compare_Dance1}), we see that the average improvement for datasets Dance1 and Odzemok is around $54\%$ (we notice that such improvements depend on the number of pixels contained in the foreground areas). %{\color{red}
Inferring from all the results, we argue that our proposed pipeline (as shown in Fig.~\ref{fig:section4-1} and Fig.~\ref{fig:new-pipeline}) can significantly lower down the processing latency at the cost of very limited quality degradation.
%} 
Therefore, such improvement can greatly help latency-sensitive applications whose quick turnaround time and high quality results drive the success of the underlying disaster response mission.

\begin{table}[t]
\caption{Latency comparison between original and optimized pipelines}
%\vspace*{-6pt}
\centering
\begin{footnotesize}
\begin{tabular}{|c||c|c|c|c|}
\hline
%\bfseries 
{\bf Step} & %\bfseries 
\begin{tabular}{@{}c@{}}{\bf Original} \\Dance1  \end{tabular}
& %\bfseries 
\begin{tabular}{@{}c@{}}{\bf Optimal}\\Dance1 \end{tabular}
&
\begin{tabular}{@{}c@{}}{\bf Original} \\Odzemok  \end{tabular}
& %\bfseries 
\begin{tabular}{@{}c@{}}{\bf Optimal}\\Odzemok \end{tabular}
%& \bfseries 
%\begin{tabular}{@{}c@{}}Original   \\Pipeline\end{tabular} 
\\
\hline\hline
%\begin{tabular}{@{}c@{}}Background\\ subtraction \end{tabular}& - & - & - & -\\
%\hline
\begin{tabular}{@{}c@{}}SfM Pipeline  \\(optimized cfgs)\end{tabular} & - & 2.94 s & - & 3.01 s\\
%\hline
%\begin{tabular}{@{}c@{}}SfM Pipeline  \\(golden cfgs)\end{tabular} & -  & - & - & -\\
\hline
\begin{tabular}{@{}c@{}}MVS Pipeline \\of FG\end{tabular}  & - & 8.78 s & - & 5.67 s\\
\hline
Others (Split, Merge)  & - & 0.26 s & - & 0.21 s\\
\hline
Total Time & {\bf 26.17} s & \begin{tabular}{@{}c@{}}\textbf{11.98} s\\(54.22\%)\end{tabular} & {\bf 19.43} s& \begin{tabular}{@{}c@{}}\textbf{8.89} s \\($54.25\%)$ \end{tabular}\\
\hline
\end{tabular}
\vspace*{-10pt}
\label{golen_compare_Dance1}
\end{footnotesize}
\end{table}

\section{Conclusions and Future Work}
\label{Sec:conclusions}
%\vspace{-0.05in}
%{\color{red}Recent techniques are using deep learning-based system for 3D reconstruction which requires which is an important development in the field. However, we didn’t consider this system in these papers because they require training samples to 3D reconstruct objects which has several problems. [Soumya will write this paragraph]}

In this paper, we developed a collaborative MEC architecture for disaster response based on an optimized framework  %optimized openMVG/openMVS pipeline 
that balances the trade-off between 3D reconstruction processing latency and quality. By exploiting the data and task level parallelism, our optimized edge-supported framework achieves a significant reduction in end-to-end latency, with negligible loss in reconstruction quality. In the future, we would like to extend this work on multiple fronts. One of the key limitations of our work is that the overall latency is still in tens of seconds when implemented on low cost compute devices. However, such latencies might not be acceptable for some real-world use cases (also restricted by use of limited CPU/GPU resources) that require 
%applications, however, there are other applications such as self-driving cars and robotic industrial automation, where the latency requirements are almost 
real-time (i.e., $<$ 100 ms) response. Therefore, one of our future research directions is to reduce the end-to-end latency %of the pipeline
even further using algorithmic optimization that uses deep learning based methods and models.
%}
%} %Additionally, we plan to develop an experimental laboratory prototype to demonstrate the efficacy of our optimization on real use cases. 
%Deep learning based 3D reconstruction methods have opened up newer avenues, and we intend to explore how to combine the benefits of traditional and neural network based methods to further improve the quality and latency of 3D reconstruction.   
%{\color{red}This is a quick and dirty conclusion, so please expand. Soumya: I have updated my parts. Saptarshi, please take a final look.}

\vspace{-0.1in}

\bibliographystyle{IEEEtran}
\bibliography{sample-base}

\end{document}